# Heterogeneous Hand Guise Classification Based on Surface Electromyographic Signals Using Multichannel Convolutional Neural Network

Niloy Sikder
*Computer Science & Engineering Discipline*
*Khulna University*
Khulna, Bangladesh
niloysikder333@gmail.com

Abu Shamim Mohammad Arif
*Computer Science & Engineering Discipline*
*Khulna University*
Khulna, Bangladesh
shamimarif@yahoo.com

Abdullah-Al Nahid
*Electronics & Communication Engineering Discipline*
*Khulna University*
Khulna, Bangladesh
nahid.ece.ku@gmail.com

***Abstract*—Electromyography (EMG) is a way of measuring the bioelectric activities that take place inside the muscles. EMG is usually performed to detect abnormalities within the nerves or muscles of a target area. The recent developments in the field of Machine Learning allow us to use EMG signals to teach machines the complex properties of human movements. Modern machines are capable of detecting numerous human activities and distinguishing among them solely based on the EMG signals produced by those activities. However, success in accomplishing this task mostly depends on the learning technique used by the machine to analyze EMG signals; and even the latest algorithms do not result in flawless classification. In this study, a novel classification method has been described employing a multichannel Convolutional Neural Network (CNN) that interprets surface EMG signals by the properties they exhibit in the power domain. The proposed method was tested on a well-established EMG dataset, and the result yields very high classification accuracy. This learning model will help researchers to develop prosthetic arms capable of detecting various hand gestures to mimic them afterwards.**

## I. Introduction

Electromyography (EMG) is a diagnostic procedure that is used to measure and document electrical activities produced by the skeletal muscles of a human body. Instrument used to carry out the procedure is called Electromyograph, and the resultant signal is called an Electromyogram. French scientist Étienne-Jules Marey recorded this voluntary muscle activity in 1890 and coined the term "Electromyography." Although EMGs are being studied for more than a century, major breakthroughs occurred within the last few decades, which helped us to develop new applications based on this technique [1]. Clinically, EMG is performed to determine the presence of nerve dysfunction, muscle dysfunction, lack of biomechanical and motor control, disruption in neuromuscular signal transmission, or other functional disorders in a specific part of the body. EMG is also useful in diagnosing various types of chronic pains (e.g., headache, low back pain) [2]. However, with the help of new Machine Learning (ML) techniques, now EMG signals are being used to detect and understand human activities or movements. The acquired knowledge are being implemented inside robots or mechanical limbs to replicate those activities. Currently, EMG is being studied and applied beyond the domains of medicine, such as Engineering, Prosthesis, and Cybernetics. Capturing the electrical activities that occur within human muscles is providing us vital information on how the muscular system works, and how to fix or replace parts of it if necessary.

An EMG signal is a collective form of multiple motor unit action potentials superimposed upon each other. The resulting signal is stochastic in nature and usually described in terms of amplitude, frequency, and phase as a function of time [3]. EMG signals can be broadly categorized based on how they are collected from the body; signals can be collected by placing electrodes on the skin and stimulating the intended nerves through small electrical impulses, or by inserting needles into a particular muscle. The former process is called surface EMG (sEMG), which was also known as Nerve Conduction studies (NCS) in the past [1]. The latter one does not require any external stimulation and collects the action potentials directly from the muscle. However, inserting needles inside the body is a painful procedure, and thus, inconvenient to perform on subjects on a regular basis. That is why Robotics and ML studies mostly use sEMG signals to study the movements of specific parts of the human body. Regardless of the background, EMG has played a vital role in the machines' understanding of limb movements. However, machines are still not capable of interpreting all types of human movements based on their sEMG signals, which provides researchers the opportunity to develop new and more potent methods for the task and advance this area of research.

The first item required to teach machines the properties of diverse limb movements based on their corresponding EMG signals is a set of noise-free, meticulously collected, and accurately labeled EMG signals. Several Universities, medical institutes, and research centers provide such datasets containing EMG signals generated from several preordained muscle movements or human activities. Among the latest publicly available sEMG datasets, one is hosted by the University of California Irvine (UCI) in their Machine Learning Repository; we will refer to it as the UCI sEMG HM dataset [4]. The researchers introduced the dataset through a conference proceeding in 2013, where they described the procedure of data collection and signal preparation. They decomposed the raw EMG signals into Intrinsic Mode Functions (IMFs) to extract features, used Principal Component Analysis (PCA) and Relief to select the best features among them, and then employed a linear classifier to carry out the classification [5]. Shortly after that, the authors reported an improvement in the classification accuracy by using Empirical Mode Decomposition (EMD) to decompose the EMG signals [6]. So far, numerous studies have been conducted on the UCI sEMG HM dataset by various groups of researchers, which resulted in many independent methods involving different feature extraction, feature selection, and classification techniques. In 2015, Ruangpaisarn and Jaiyen proposed a method incorporating Singular Value Decomposition (SVD) for feature extraction, and Sequential Minimal Optimization (SMO) for classification of the six different hand movement signals [7]. In 2017, Kim and Pan published an article describing a method that uses non-uniform filter bank and Waveform Length to extract features, PCA and Linear Discriminant Analysis (LDA) to select the most significant features, and Euclidean Distance (ED) and Support Vector Machine (SVM) and $k$-Nearest Neighbors ($k$-NN) for classification [8]. Later that year, Iqbal, Fattah and Zahin used SVD and PCA for feature extraction and selection, and then $k$-NN for classifying the figure position classes [9]. In 2019, Ramírez-Martínez et al. used Burg reflection coefficients to build an optimal feature set, and carried out a series of classification tasks using different classifiers and preprocessing techniques [10]. In the same year, Nishad et al. designed a cost-effective sEMG classification method applying Tunable-Q Wavelet Transform (TQWT) based filter-bank for signal decomposition, Kraskov Entropy (KRE) for feature extraction, and finally, $k$-NN for carrying out the classification of various hand movements [11].

Although the use of deep learning models is less common in hand gesture classification, such approaches are not rare. For example, in 2019, Wei et al. described a gesture recognition model where they used a Convolutional Neural network (CNN) framework that operated on sEMG images constructed from sEMG signals collected form three sEMG databases [12]. In the previous year, Hu et al. provided a hybrid CNN-RNN architecture for the same purpose and tested it on five different sEMG datasets [13]. In this paper, we will describe a novel approach for hand guise classification using a multichannel-CNN model that works depending on the power information of their corresponding sEMG signals.

The rest of the paper is organized as follows. Section II describes the methodology of the proposed algorithm with brief discussions on the UCI sEMG HM dataset, the feature extraction method, and the fundamentals of CNN. Section III narrates the results acquired using this method, evaluates its performance based on some well-known parameters, and compares them with some of the state-of-the-art methods described in similar studies. Finally, Section IV summarizes the paper and imparts some scopes for further researches.

## II. Methodology

This study aims to classify the six different hand positions based on their sEMG signals contained by the UCI sEMG HM dataset with the help of a two-channel CNN model. Fig. 1 illustrates the principal methodology of the proposed algorithm. This is a supervised learning algorithm, so it involves a training stage where the model is trained with practical sEMG signals or their corresponding features, and a subsequent test stage where the knowledge acquired by the model in the previous stage is put into test. The following sub- sections provide a detailed look at the key components of the method before moving to the experimental phase.

### *A. UCI sEMG HM Dataset*

The UCI sEMG HM dataset contains sEMG signals collected from six different subjects while holding objects of different size and shape. Three of the subjects are female and the rest three are male – all between the age bracket of 20 and 22 years. The repository has two separate datasets. We will refer to them as DataBase1 and DataBase2. The former one contains EMG samples collected from five different subjects (three females and two males), and the latter one contains samples provided by a single male subject in three different days. Table I provides a graphical representation of the objects that were griped, the number of samples in both datasets, and the corresponding class labels. The signals were collected using a two-channel electrode at a sampling rate of 500 Hz [5]. Prior to including in the datasets, the signals were filtered using a Butterworth Band-pass filter with low and high cutoff frequency of 15 Hz and 500 Hz respectively, and a Notch filter to eliminate line interference.

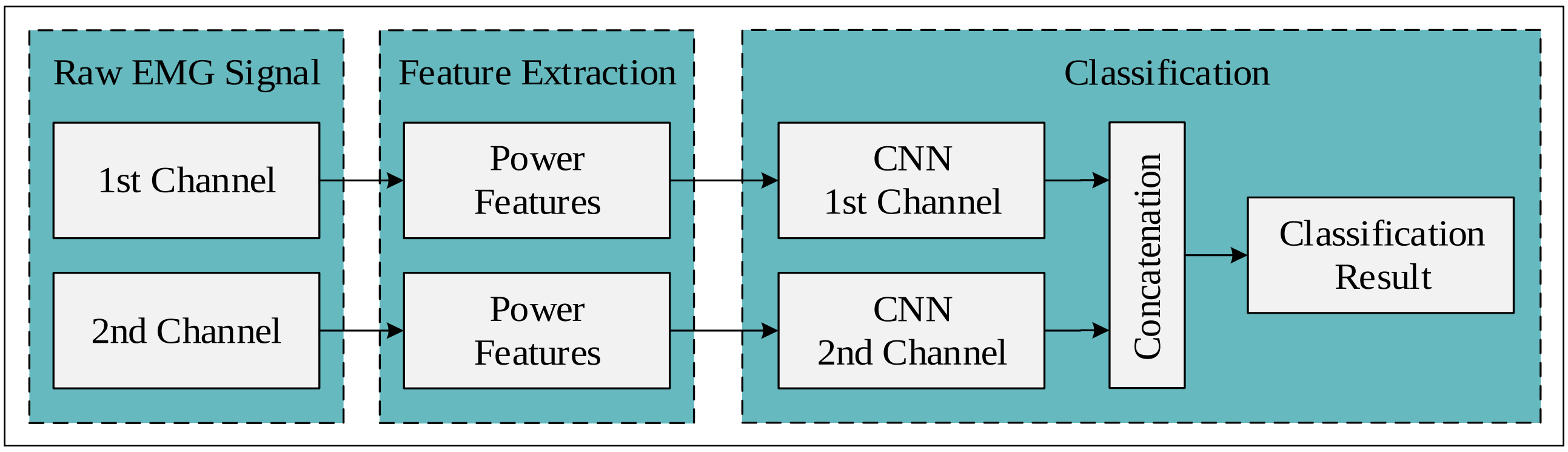

Fig. 1. The Proposed CNN-based sEMG signal classification model.

TABLE I. MOVEMENTS, THEIR LABELS, AND NUMBER OF SAMPLES

| **Action** | **Object Type** | **Class Label** | **Samples in DataBase1** | **Samples in DataBase2** |
|---|---|---|---|---|
| 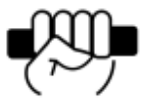 | Cylindrical | C | 30×5=150 | 100×3=300 |
| 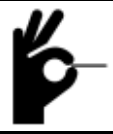 | Tip | T | 30×5=150 | 100×3=300 |
| 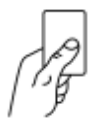 | Lateral | L | 30×5=150 | 100×3=300 |
| 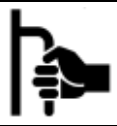 | Hook | H | 30×5=150 | 100×3=300 |
| 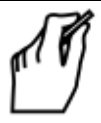 | Palmer | P | 30×5=150 | 100×3=300 |
| 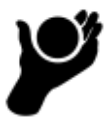 | Spherical | S | 30×5=150 | 100×3=300 |
| Total | | | 900 | 1800 |

Fig. 2 shows the t-distributed stochastic neighbor embedding (t-SNE) graph of the samples present in the databases. The graphs signify how the samples of different classes are distributed in a 2D space. As we can see, samples are all scattered across the plane, which is not an ideal condition for any classifier to operate. The samples of the same class should be clustered together as densely as possible, whereas the clusters of different classes should as far away as possible from each other. That is why meaningful features need to be extracted from the samples to provide the classifier distinguishable characteristics to carry out its operation.

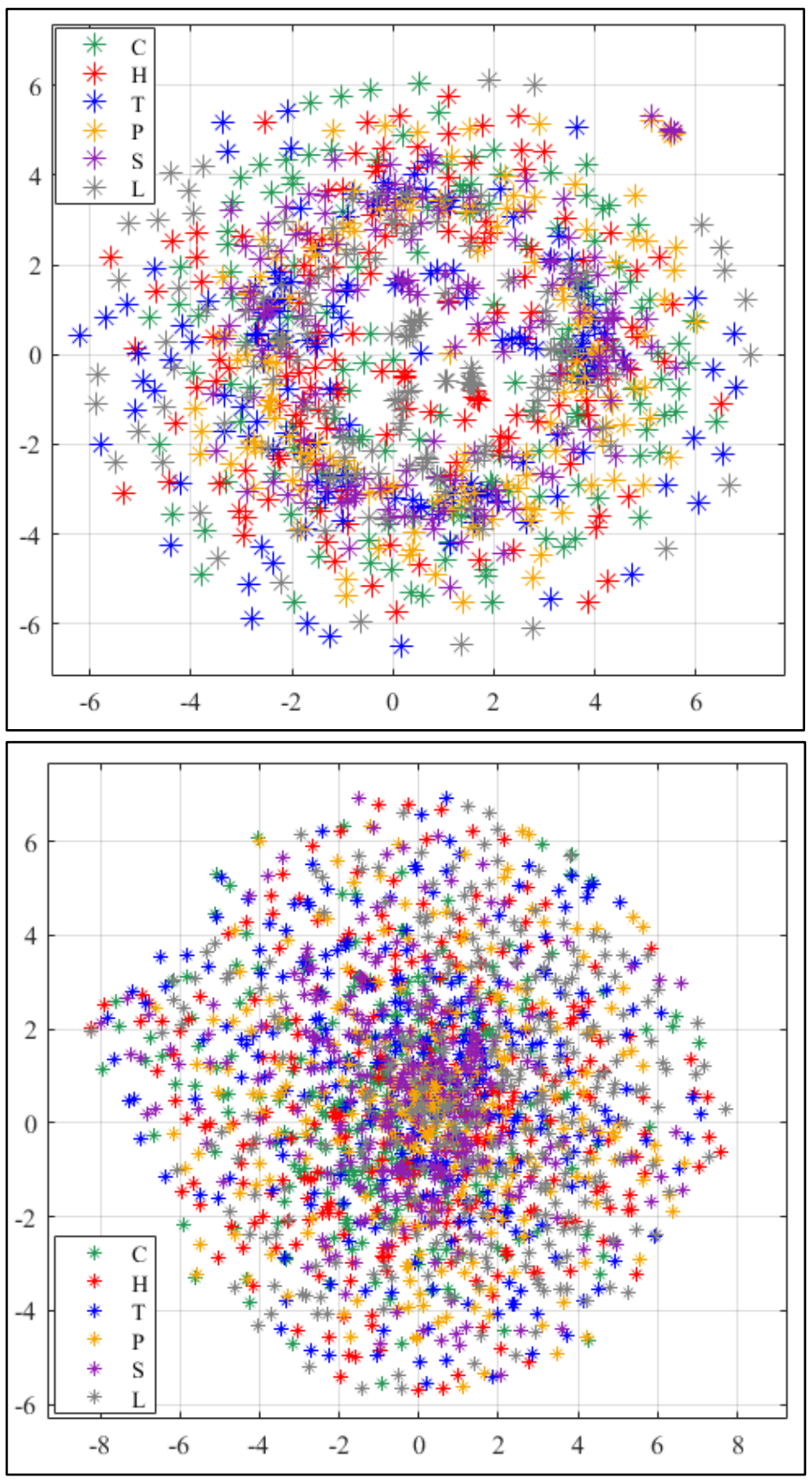

(a)

(b)

Fig. 2. t-SNE of the sEMG samples of (a) DataBase1, (b) DataBase2.

### B. *Power Feature Extraction Using Burg's Method*

The power spectral density (PSD) of a signal illustrates how much power the signal possess at different frequencies. There are many methods available to calculate the PSD of a time-variant signal. The algorithm we used in this study was developed by Günter Burg in 1975, also known as Burg's autoregressive (AR) model [14]. A model that predicts its future outcomes based on the outcomes it encountered in the past is referred to as an AR model. The Burg's algorithm is a technique to fit an AR model into a time-variant signal. The method is different from auto-correlation, covariance, and Fourier-based methods because it does not make any assumptions on the signal values in advance [15]. The algorithm characterizes the AR model in terms of reflection coefficients ($R_i$) for the equivalent lattice filter implementation. The coefficients are computed in a sequential manner using a version of the Levinson Recursion. These coefficients minimize the sum of the forward and backward prediction error powers at each stage. If we take a time-series signal $\boldsymbol{x} = \{x_0, x_1, \dots, x_{N-1}\}$, the Burg algorithm computes the $R_i$ for $x_i$ using:

$$R_i = \frac{\sum_{n=i}^{N-1}\{e_{i-1}^{f}[n] \times e_{i-1}^{b^*}[n-1]\}}{\sum_{n=i}^{N-1}\left\{\left|e_{i-1}^{f}[n]\right|^2 + \left|e_{i-1}^{b}[n-1]\right|^2\right\}} \qquad (1)$$

where $e_i^f[n]$ is the forward estimation error and $e_i^b[n]$ is the backward estimation error of the $i$th iteration. These errors are defined as:

$$e_i^f[n] = e_{i-1}^f[n] + R_i e_{i-1}^b[n-1] \qquad (2)$$

and,

$$e_i^b[n] = e_{i-1}^b[n-1] + R_i^* e_{i-1}^f[n] \qquad (3)$$

If values of $R_1, R_2, \dots, R_{i-1}$ are selected and remain fixed, the stage $i$ error for $R_i$ can be calculated using (1):

$$\mathcal{E}_i = \sum_{n=i}^{N-1}\left|e_{i-1}^f[n]\right|^2 + \sum_{n=i}^{N-1}\left|e_{i-1}^b[n]\right|^2 \qquad (4)$$

Now, the AR filter coefficients can be obtained recursively as:

$$\mathcal{A}_i[j] = \begin{cases} \mathcal{A}_{i-1}[j] + R_i \mathcal{A}_{i-1}^*[j-1], & when\ \ j < i \\ R_i \qquad , & when\ \ j = i \end{cases} \qquad (5)$$

where $i$ is the order of the model, and $j$ is the delay index of the specified coefficient. The Burg's method is relatively simple, and compliant to both order-recursive and time-recursive solutions. Fig. 3(a) presents the 10$^{th}$ order AR power spectral density estimate of a sEMG signal using the Burg's method. Fig. 3(b) and 3(c) show that the samples of 2(a) and 2(b) have become more clustered if their power features are taken into account.

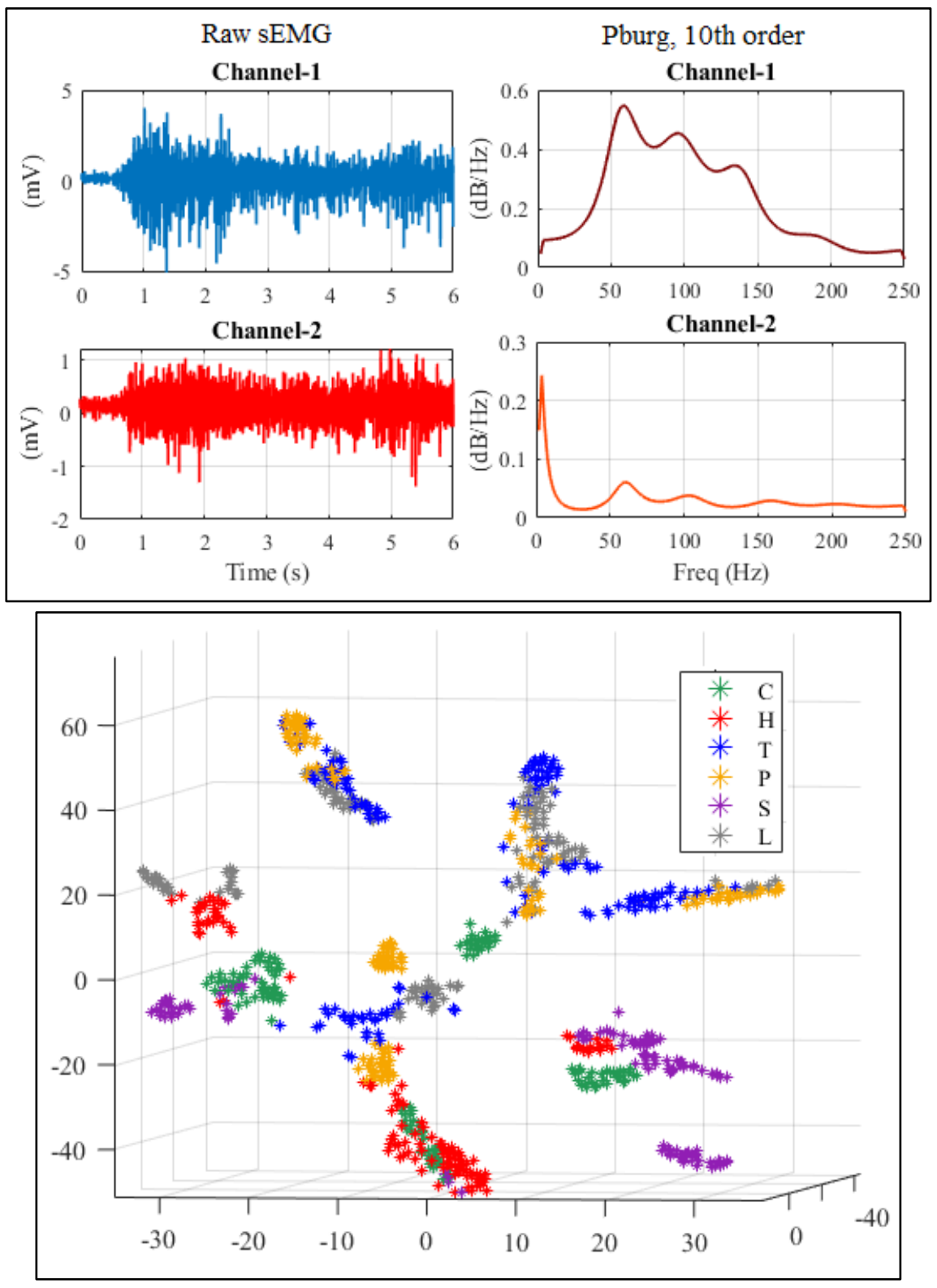

(a)

(b)

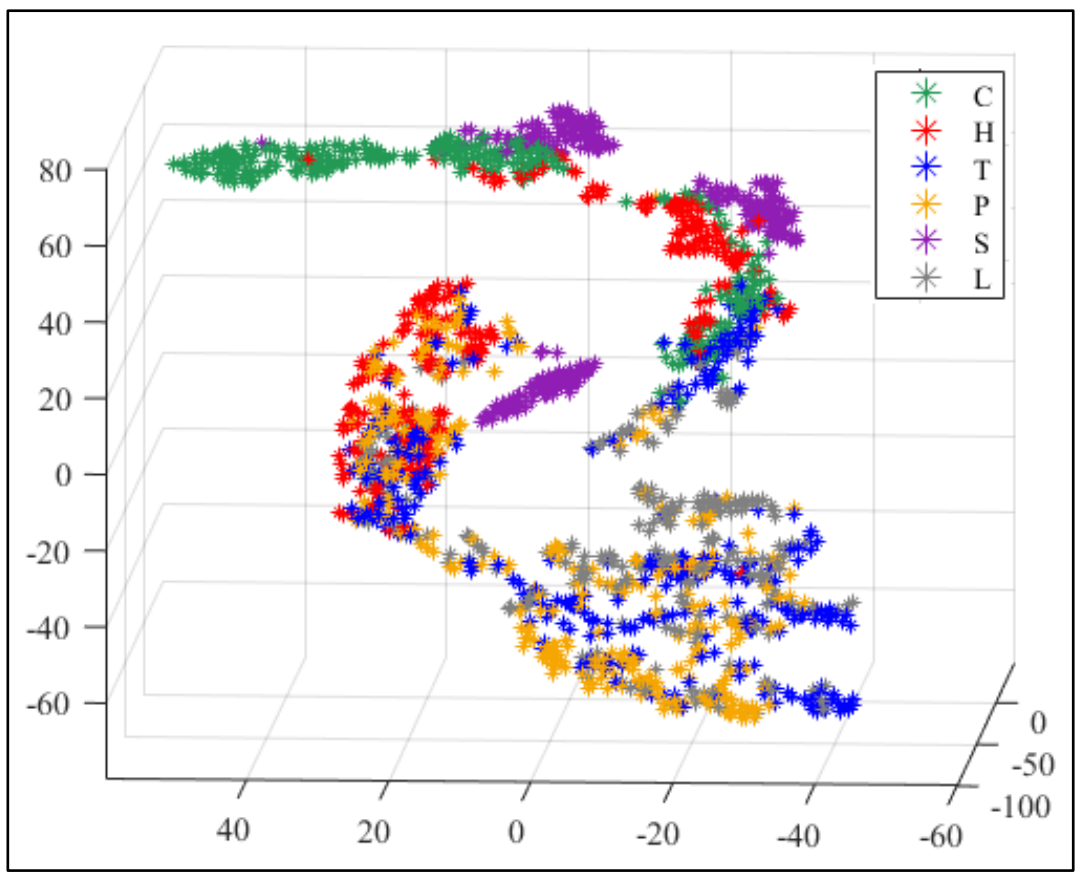

(c)

Fig. 3. (a) AR 10th order power spectral density estimate of a 2-channel sEMG signal using Burg's method. And t-SNE of the sEMG samples of (b) DataBase1, and (c) DataBase2 based on their PSD.

### C. Multichannel CNN for Hand Movement Classification

The fundamental idea behind the architecture of CNN has come from biological neurons. CNNs can process data that has a grid-like topology. At the heart of CNN, there is a mathematical operation that slides one function over another and measures the integral of their point-wise multiplication, commonly known as convolution [16]. A neural network that uses convolution instead of general matrix multiplication in at least one layer is called a convolutional network [17]. CNN was developed based on Hubel and Wiesel's explanation of how a cat's visual cortex works, where they observed that specific sections of the visual field stimulate a set of particular neurons [18]. Although CCN primarily works high-dimensional images, they are also effective on low-dimensional signals. CNNs have been widely used in audio processing, speech recognition, machine fault classification, and different kinds of biomedical signal classification such as Electroencephalogram (EEG), Electrocardiogram (ECG), and Human Activity Recognition (HAR). As stated earlier, in the model, we are integrating two different CNNs (specified as channels). As shown in Fig. 4, the first channel of CNN will process the power features extracted from the sEMG signals collected through the first electrode, and similarly, the second one will process the features extracted from the signals collected through the second electrode. The basic parameters for both the channels are the same.

To anticipate the working procedure of each unit of CNN, let us consider a sequence of $M$-dimensional observations which are arranged in a mini-batch having length $L$, the output of a hidden layer $\ell$ at node $d$ and time $t$ can be calculated from the input $x_t \in \mathcal{R}^M$ as:

$$\ell_{d,t} = \sigma\left(\sum_{i=0}^{M-1} \omega_{d,i}\, x_{i,t} + b_d\right) \quad (6)$$

where $\sigma$ is a non-linear function, $t$ lies between 0 and $L$, $\omega_{d,i}$ are the corresponding weights, and $b_d$ is the bias [19]. A convolutional layer arranges the weights as a two-dimensional patch whose parameters depend on $d$. The output of such a layer at different points of the matrix $(s,t)$ will be:

$$\ell_{d,s,t} = \sigma\left[\sum_{i=s}^{s+n-1}\left(\sum_{j=t}^{t+n'-1} \omega_{d,i-s,j-t}\, x_{i,j} + b_d\right)\right] \tag{7}$$

where the weight patch $\omega \in \mathcal{R}^{n\times n'}(n < M, n' < L)$ is multiplied by the parts of the input having direct neighborhood of the position $(s,t)$. The set of outputs of $d$, also known as the feature map, are analogous to a feature stream extracted by $d$. We are suppressing the details on patch symmetry and boundary handling at this point. However, within the training framework, it is advisable to have a feature map with the same number of columns of the mini-batch [19]. Now, the output of each filter of the first layer is fed as the input of the following layer in CNN. If there are $N$ input streams arranged in mini-batches of the same dimension, the corresponding output can be calculated using (7):

$$\ell_{d,s,t} = \sigma\left[\sum_{i=s}^{s+n-1}\left\{\sum_{j=t}^{t+n'-1}\left(\sum_{r=0}^{N-1} \omega_{d,i-s,j-t,r} x_{i,j,r} + b_d\right)\right\}\right] \tag{8}$$

In the case of the power features collected from sEMG signals, the output of a convolutional unit $d$ would be:

$$\ell_{d,t} = \sigma\left(\sum_{i=s}^{p+n-1} \omega_{d,i-s}\, x_i + b_d\right) \tag{9}$$

where $\omega \in \mathcal{R}^{n}$ is a weight vector, and $p$ is the position within the output vector. The step size, denoted by $z$, is chosen in such a way that the output feature stream is only calculated for positions defined by $\{\ell_{d,s\cdot z} : 0 \le s \cdot z < M - n\}$. The outputs of the dense layer of the two channels were then concatenated, and the “Softmax” activation function was used to perform the final sEMG classification.

## III. Results and Discussion

In the previous sections, we have discussed the hand movement signals of the UCI sEMG HM dataset, and our strategy to classify them using a two-channel CNN model. In this section, we will present our findings. In accordance with the described procedure, we set a classification model where 70% samples (630 for DataBase1 and 1260 for DataBase2) were used to train the two-channel CNN model, and the rest of the samples were used to test it. Fig. 5 delineates the outcome of the classification. As we can see, the model can classify the hand movement signals of DataBase1 with a 96.3% accuracy at the end of 300 epochs, which can reach up to 98.5%. As for DataBase2, the model achieved an accuracy of 88.9% with a peak of 90.56%. Fig. 6 also points out that the model passed the 90% mark for DataBase1 and the 80% mark for DataBase2 within the first 20 epochs and rarely fell afterward in both cases, which indicates that the model is very stable. Moreover, Fig. 5 provides information regarding the loss in each epoch of classification. It can be easily anticipated that the accuracy decreased whenever the loss increased. For DataBase2, the loss values were unpredictable, which affected the model's performance and restricted the accuracy scores from reaching the 91% mark.

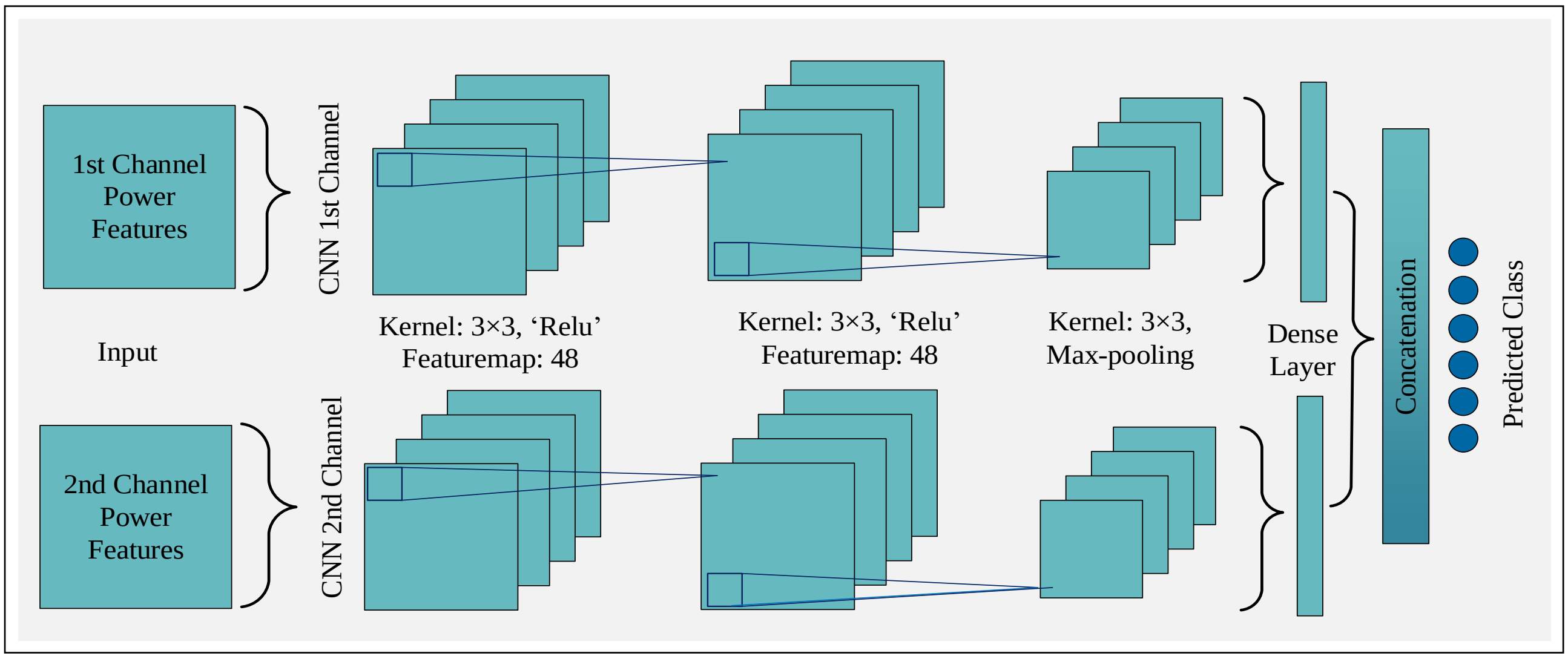

Fig. 4. Proposed two-channel CNN for sEMG classification

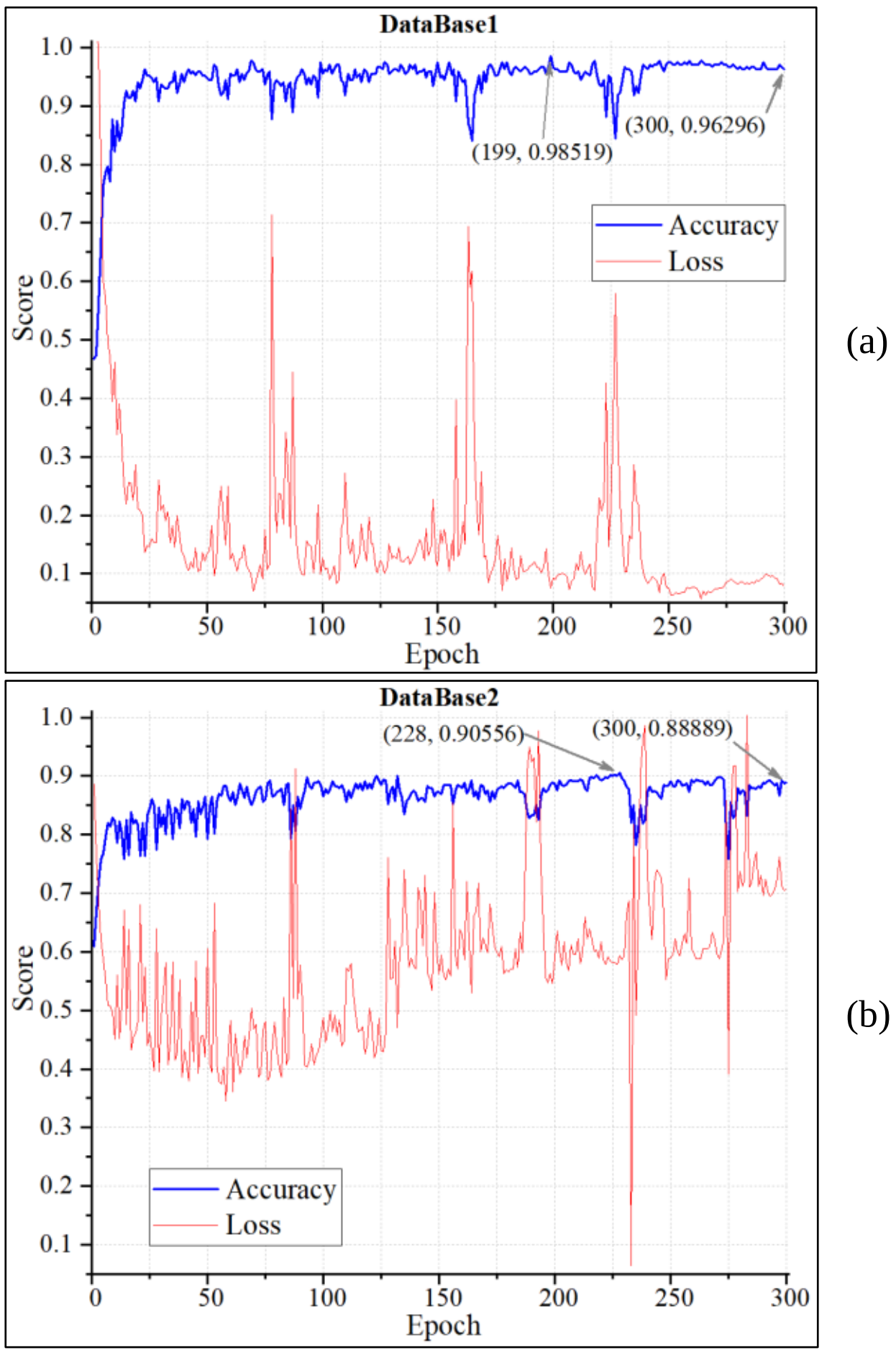

Fig. 5. Accuracy and loss at each epoch while classifying the samples of (a) DataBase1, (b) DataBase2.

Being inquisitive about the performance of our model, we tested it on the individual sets of data present in DataBase1 and DataBase2 as well. As mentioned in Section II, the first database is comprised of hand movement samples collected from five different subjects and the second one accommodates the data of a single subject collected in three different days. If we consider the samples of each subject and day discretely, then we have eight different subsets of hand movement data. Fig. 6 presents the performance of our model on each of these subsets along with their collective form (DataBase1 and DataBase2) in terms of the model accuracy (the classification accuracy of the 300$^{th}$ epoch), the maximum accuracy (the peak value of accuracy reached within the 300 epochs) and the average accuracy (an algebraic mean of all the accuracy scores). The figure reveals that the model could classify the samples provided by the male subjects in DataBase1 with almost the perfect accuracy. However, the performance of the model declined while deciding the labels of the other samples provided by the female participates. And conjointly, the model achieved a 96.3% accuracy at the last epoch of the classification operation on DataBase1.

In the case of DataBase2, samples collected on the second day seemed to be more distinguishable to the model than the samples collected on the other two days achieving a peak accuracy of 96.67%. However, the other two subsets and the dataset itself were not distinguishable enough to get beyond the 92% mark. Fig. 6 also provides the corresponding F1-scores of the classification operations performed on different datasets and data subsets. The F1-score is a weighted average of the precision and recall scores of a classification operation [20]. The former score signifies the amount of the correctly labeled samples within the samples that have been labeled as a specific class, whereas the latter one reveals the fraction of the instances of a particular class that the classifier recognizes correctly. The F1-score is at times a better parameter to judge the performance of a classifier than the classification accuracy, especially while working with imbalanced datasets since it takes both the precision and recall scores into account [21]. As our model achieved F1-scores almost identical to the corresponding accuracy values, it can be said with confidence that the model can provide the claimed performance.

Furthermore, to provide an in-depth look at the performance of our model while classifying multiples hand movement classes, we present the confusion matrices of the classification operations performed on DataBase1 and DataBase2 in Fig. 7. The figure indicates that with only four samples misclassified, the classification outcome of the test samples of DataBase1 was almost flawless. In the case of DataBase2, however, the model wrongly labeled several samples, which is also reflected in the results provided in Fig. 5(b) and Fig. 6. The presented confusion matrices contain the classification information of the iterations when the maximum accuracies were achieved (i.e., 199th epoch of DataBase1 and 228th epoch of DataBase2).

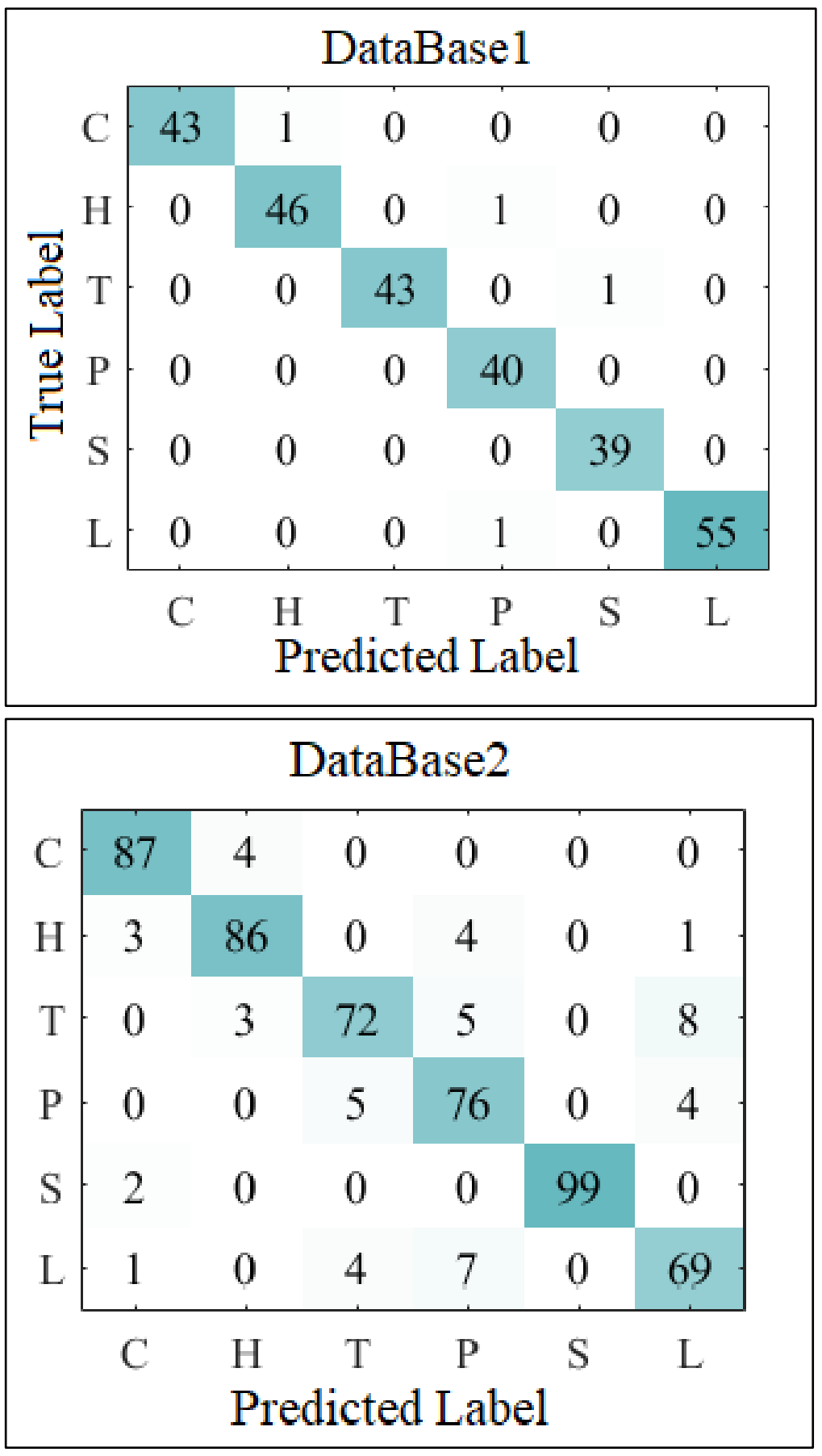

(a)

(b)

Fig. 7. Confusion matrix of the classification operation using the proposed model on (a) DataBase1, and (b) DataBase2.

To put a perspective on the performance of our model, we have compared the obtained results with four other similar studies which involve the UCI sEMG HM dataset in Table II. As the table shows, in terms of the average classification accuracy of DataBase1, the proposed method outperforms the methods described in [6], [7] and [9] by 9.31%, 0.3%, and 11.81% respectively. Only [11] have attained better performances than the proposed model. As of data subsets, only [11] has achieved significantly better classification accuracy than our model on the Female3 subset. However, if the maximum accuracy scores are considered, the proposed model outperforms the models described in those references by some margins.

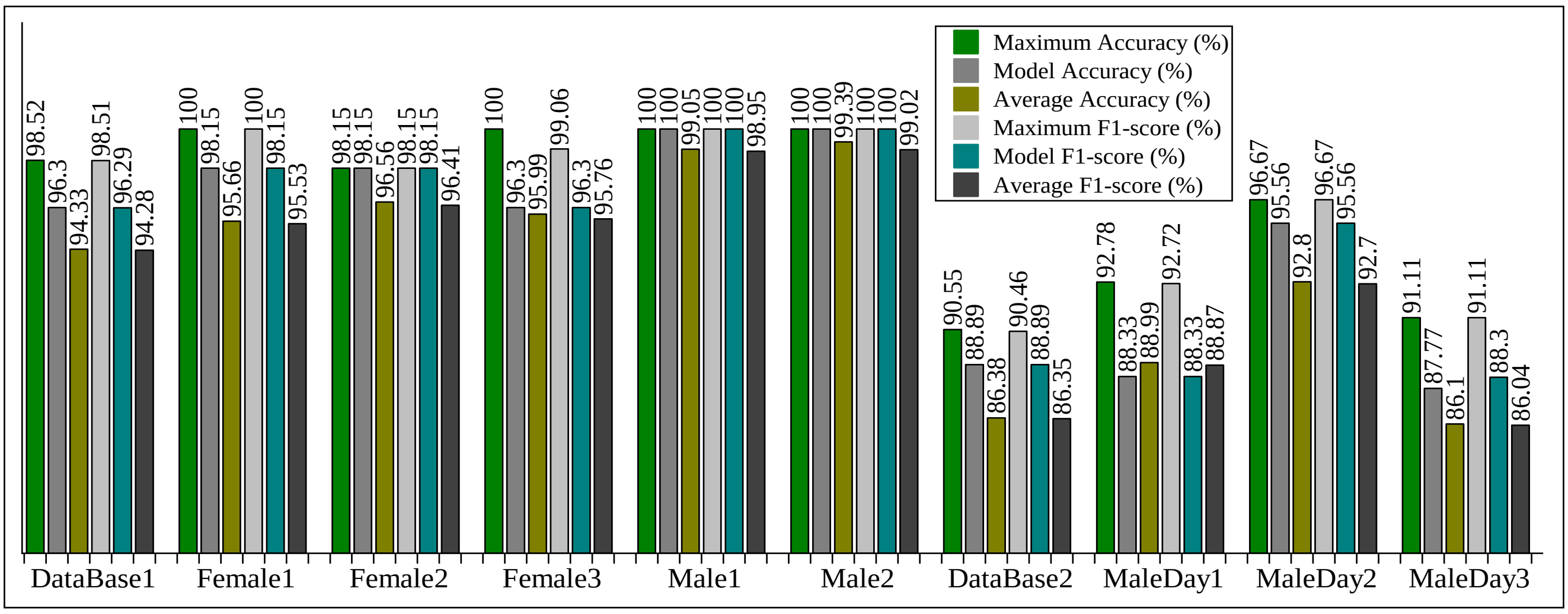

Fig. 6. Performance of the proposed model on various datasets and their subsets.

TABLE II. PERFORMANCE COMPARISON WITH OTHER STUDIES

| **Subject** | **Accuracy (%)** | | | | | |
|---|---|---|---|---|---|---|
| | Proposed | | [6] | [7] | [9] | [11] |
| | Max | Model | | | | |
| Female1 | 100 | 98.15 | 87.25 | 96.67 | 82.78 | 98.33 |
| Female2 | 98.15 | 98.15 | 88.05 | 98.89 | 87.67 | 97.78 |
| Female3 | 100 | 96.3 | 85.53 | 96.67 | 83.11 | 99.44 |
| Male1 | 100 | 100 | 90.42 | 98.89 | 90 | 98.89 |
| Male2 | 100 | 100 | 94.8 | 100 | 90 | 98.33 |
| **Average** | **99.63** | **98.52** | **89.21** | **98.22** | **86.71** | **98.55** |

## IV. CONCLUSIONS

The paper described a classification model for different hand movements, tested it on the UCI sEMG HM dataset, and presented the obtained outcomes. The results yield a 96.3% model classification accuracy on the first database and 88.89% on the second one. However, adding more nodes and layers in the CNN architecture may improve the classification performance. A different set of features can also be extracted and combined with the existing ones to make the signals more distinguishable. Although the model works very well on DataBase1, DataBase2 still poses challenges, solving these difficulties is subjected to future studies.